%% file: main.tex
\documentclass{article}
\usepackage[preprint]{neurips_2024}

\usepackage[utf8]{inputenc} 
\usepackage[T1]{fontenc}    
\usepackage{hyperref}       
\usepackage{url}            
\usepackage{booktabs}       
\usepackage{amsfonts}       
\usepackage{nicefrac}       
\usepackage{microtype}      
\usepackage{xcolor}         
 \usepackage{amsmath} 
\usepackage{graphicx}
\usepackage{makecell}
\usepackage{subcaption} 
\usepackage{algorithm}
\usepackage{algorithmic}
\usepackage{wrapfig}
\usepackage{caption}
\usepackage{subcaption}

\title{Are Long-LLMs A Necessity For Long-Context Tasks?}

\author{Hongjin Qian$^{1,2}$,~Zheng Liu$^2$\thanks{Corresponding author.},~Peitian Zhang$^1$,~Kelong Mao$^1$,~Yujia Zhou$^1$\\
\textbf{Xu Chen}$^1$,~\textbf{Zhicheng Dou}$^1$\\
        $^1$ Gaoling School of Artificial Intelligence, Renmin University of China \\ 
        $^2$ Beijing Academy of Artificial Intelligence \\
        \texttt{\{chienqhj,zhengliu1026\}@gmail.com} \\}

\include{math_command}

\begin{document}

\maketitle

\begin{abstract}

    The learning and deployment of long-LLMs remains a challenging problem despite recent progresses. In this work, we argue that the long-LLMs are not a necessity to solve long-context tasks, as common long-context tasks are short-context solvable, i.e. they can be solved by purely working with oracle short-contexts within the long-context tasks' inputs. On top of this argument, we propose a framework called \textbf{LC-Boost} (\underline{L}ong-\underline{C}ontext \underline{Boo}t\underline{st}rapper), which enables a short-LLM to address the long-context tasks in a bootstrapping manner. In our framework, the short-LLM prompts itself to reason for two critical decisions: 1) how to access to the appropriate part of context within the input, 2) how to make effective use of the accessed context. By adaptively accessing and utilizing the context based on the presented tasks, LC-Boost can serve as a general framework to handle diversified long-context processing problems. We comprehensively evaluate different types of tasks from popular long-context benchmarks, where LC-Boost is able to achieve a substantially improved performance with a much smaller consumption of resource.   
\end{abstract}

\section{Introduction}
\label{sec:intro}
Large language models (LLMs) are widely adopted for real-world applications. Many of the applications are associated with long-sequence inputs, such as long-document question answering and summarization. As such, the LLMs are commonly expected to have a long working context (\textit{a.k.a.} long-LLMs) in order to confront such demanding scenarios~\citep{bai2023longbench,zhang2024inftybench}. Unfortunately, the learning and deployment of long-LLMs are still challenging in multiple perspectives. Particularly, many existing LLMs are initially introduced with a limited size of context (\eg, 2K for Llama-1 \cite{touvron2023llama}, 4K for Llama-2 \cite{llama}, 8K for Llama-3~\footnote{\url{https://llama.meta.com/llama3/}}). Although the initial short-LLM can be fine-tuned to establish a much longer context, it is likely to take substantial costs; and more seriously, it is extremely resource-consuming to deploy the long-LLMs~\citep{kaplan2020scaling}. The continually training may also compromise the LLMs' general capability over short contexts~\citep{liu2023lost,li2023long}. In fact, it remains an open problem to explore new solutions which may tackle long-context tasks both effectively and efficiently. 

In this paper, we argue that most long-context tasks are \textit{short-context solvable}. That is to say, the long-context tasks, despite associated with long-sequence inputs, can be addressed by merely working with short-contexts in a strategic way. For example, the reading comprehension or summarization of a book can be solved based on the extraction of necessary key facts from the book.
The above argument is akin to the working patterns of human beings and modern computers, where arbitrary long-form problems can always be decomposed and solved on top of a limited memory capacity~\citep{adolphs1999social,bryant2011computer}. 
However, even if the above argument holds, it is still non-trivial to solve the long-context tasks purely based on short contexts. This is because different tasks call for distinct ways of accessing and utilizing information from the long context; therefore, there can hardly be any fixed rules to handle all possible situations. To address this challenge, we propose a method, called \textbf{LC-Boost}, where short-LLMs are employed to solve general long-context tasks in a bootstrapping manner. LC-Boost operates with two critical reasoning steps. One is the reasoning of \textbf{Access}, where the LLM prompts itself to plan for how to access the appropriate part of context within the input. The other one is the reasoning of \textbf{Utilize}, where the LLM figures out how to make effective use of the accessed context. Thanks to the above design, LC-Boost is able to adaptively handle diversified long-context tasks according to their unique nature. For example, given a knowledge-grounded QA problem, the LLM may directly access to the knowledgable context through retrieval, and generate the answer in the form of RAG. Besides, it may sequentially scan the long context chunk-by-chunk if the task calls for the aggregation of specific information from the entire input.  
\begin{figure}
    \centering
    \includegraphics[width=\linewidth]{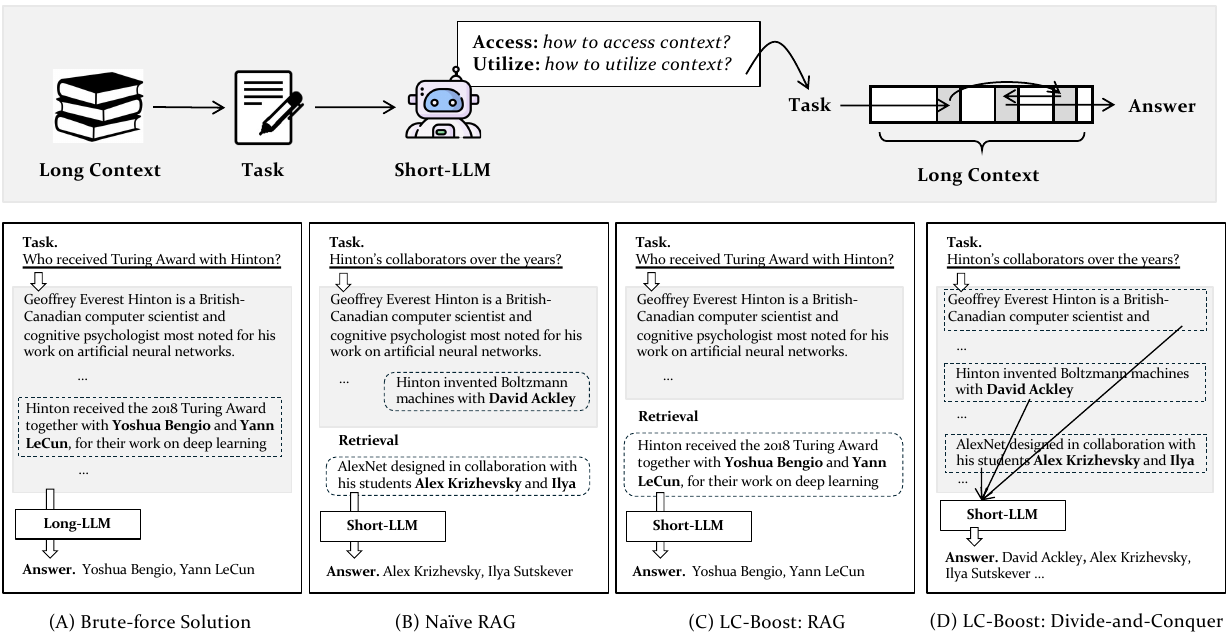}
    \caption{\textbf{Illustration for LC-Boost}. The LLM is prompted to reason for how to access to proper context and how to utilize the accessed context to solve the task. \textbf{Toy Examples.}
    (A) Brute-force solution. Despite correctness, it is unnecessarily expensive due to the processing of the entire context simultaneously. (B) Naive RAG. It is hard to handle problems like information aggregation, which leads to the incomplete answer. (C) LC-Boost leverages RAG to tackle the problem, which produces the correct answer in a small cost. (D) LC-Boost processes the long-context via sequential scan, which correctly solves the problem based on the comprehensively collected information.}  
    \label{fig:frame} 
\end{figure}

The following toy examples are presented to better illustrate the mechanism of LC-Boost (Figure \ref{fig:frame}). Particular, there are two common approaches to tackle long-context problems: (A) the brute-force method based on long-LLMs, (B) the surrogate methods, like RAG \cite{longctx}. Despite being straightforward, the brute-force method is likely to incur huge unnecessary costs as the problem could be directly solved by simple surrogate methods, like RAG. On the other hand, although the surrogate methods may help in certain cases, they are likely to become useless in other situations. For instance, the RAG-based methods are inappropriate to handle information aggregation problems, as showcased in~(B). In contrast, LC-Boost is able to handle general long-context tasks thanks to the proper reasoning of how to access and utilize
the long-context information based on each specific task. As shown in (C), it can directly access to the needed information via retrieval and generate the answer based on RAG. Meanwhile, it can also process the entire context in a divide-and-conquer manner, which will fully collect the information and solve the problem presented in (D).  

We perform comprehensive experiments for LC-Boost, including both popular real-world long-context problems, like question-answering and summarization of long documents, and a wide variety of synthetic tasks. In our experiments, LC-Boost is able to achieve equivalent performances as the brute-force methods based on strong long-LLMs, \eg, GPT-4-128K. In many cases, its performances can even notably surpass the brute-force methods, probably due to the elimination of distracting context. Besides, our experiments also underscore the importance of reasoning and adaptability, as LC-Boost outperforms all short-LLM surrogates with predefined access and utilization of context. 

To summarize, our paper makes the following contributions. (1) We identify the research problem of solving long-context problems with short-LLMs. To the best of our knowledge, it is the first study of its kind, which is important to not only address the problem itself but also meaningful to the sustainability and energy-efficient running of AI industry in a broader sense. (2) We propose a novel framework LC-Boost, which is able to adaptively handle general long-context tasks based on the reasoning of how to access and utilize the long context. (3) We empirically verify the effectiveness of LC-Boost based on its superior performances achieved from low resource-consumption.

\section{LC-Boost}

\subsection{Preliminaries}
LLMs can be succinctly defined as $\gY=\gamma(q)$, where $\gamma(\cdot)$ represents a selected LLM, $q$ denotes a user query, and $\gY$ refers to the answer produced by the LLMs. As highlighted in many previous studies, \eg,~\citep{ji2023survey,lewis2020retrieval,shuster-etal-2021-retrieval-augmentation}, the knowledge embedded in an LLM's parameters is static and, consequently, often fails to adequately address user queries requiring up-to-date or in-depth knowledge. To address this limitation, we can introduce external knowledge (refer to as context $\gX$) into the LLMs. Additionally, tasks involving information aggregation~(\eg, summarization) also take a context $\gX$ as input along with task instructions $q$. Thus, we can generally define the model's generation process w.r.t. a context $\gX$ as:
$
    \gY=\gamma(q, \gX).
$

As discussed in Section \ref{sec:intro}, in many scenarios, the context $\gX$ is a long sequence, necessitating that LLMs manage long contexts. However, most existing LLMs were originally introduced with limited context sizes~(\eg, 4K). Consequently, these models are unable to process inputs that exceed their capacity without truncation. In this paper, we characterize such scenarios as \textit{long-context problem}. It involves LLMs processing inputs that notably surpass their inherent context limitations, which can be formally described by:
\begin{equation}
    \gY=\gamma(q, \gX)\quad \text{s.t.} |\gX|\gg L,
    \label{eq:long}
\end{equation}
where \(L\) denotes the native context length limit of the LLM. The most straightforward way to address the long-context problem is to increase the LLMs' context length \(L\), mitigating the challenges of long contexts. In this paper, we instead explore solving long-context tasks using short-context LLMs~(\eg, 4K) without increasing the model's context length \(L\). 

\subsection{Pilot Study: Are Most Long-Context Tasks Short-Context Solvable?}
Despite the potential for fine-tuning LLMs to handle much longer contexts, this approach incurs substantial costs. Additionally, directly processing long contexts during the inference stage exponentially increases computing resource consumption, which is not environmentally friendly. 
In the following, we conduct a pilot study from both theoretical and empirical perspectives to explore the question: Are most long-context tasks solvable with short contexts?

\paragraph{Theoretical Analysis}
Suppose we have an input variable $\gX$ and an output variable $\gY$, the relevant part of $\gX$ given $\gY$ is denoted by $\tilde{\gX}$. An ideal $\tilde{\gX}$ should capture all relevant features of the original input variable $\gX$ in relation to $\gY$. In other words, the optimal $\tilde{\gX}$ represents the simplest mapping of $\gX$ that accurately preserves the mutual information $I(\gX; \gY)$. We therefore propose a Markov chain $\gX \rightarrow \tilde{\gX} \rightarrow \gY$. According to the data processing inequality (DPI), we have $I(\gX; \tilde{\gX}) \geq I(\gX; \gY)$, with equality holding if and only if $\tilde{\gX}$ constitutes a \textit{sufficient statistics}~\citep{cover1999elements,tishby2015deep}. This suggests that, in an optimal setting, we can always find a subset \(\tilde{\mathcal{X}} \subseteq \mathcal{X}\) that provides information at least as useful for generating the output \(\mathcal{Y}\) as the full context \(\mathcal{X}\).

In practical scenarios, obtaining the optimal $\tilde{\gX}$ is challenging due to various factors, such as empirical errors~\cite{mohri2018foundations}. Thus, we can only estimate $\tilde{\gX}$. Estimating $\tilde{\gX}$ directly from $\gX$ might be challenging if $\gX$ defines a large variable space. In this situation, we propose decomposing the original input variable $\gX$ into a series of subsets, $\gX = \{\gX_1, \cdots, \gX_n\}$ and process each subset variable separately. Thus, according to the \textit{chain rule for mutual information}~\cite{cover1999elements}, we have:
\begin{align}
    I(\gX,\tilde{\gX}) = I(\gX_1,\cdots,\gX_n;\tilde{\gX}) = I(\gX_1;\tilde{\gX})+ \sum\limits_{i=2}^n I(\gX_i; \tilde{\gX} | \gX_1, \cdots, \gX_{i-1}),
\label{eq:decomp}
\end{align}
which indicates that the mutual information $I(\gX, \tilde{\gX})$ can be understood as the sum of the mutual information of each subset $\gX_i$ and $\tilde{\gX}_i$ given all previous subsets.

In the scenario of Eq.~\ref{eq:long}, the variable \(\mathcal{X}\) represents a long context and the variable \(\mathcal{Y}\) is the output answer produced by a LLM. Thus, $\tilde{\gX}$ can be interpreted as the \textit{minimal necessary context} from the long context $\gX$ given the output answer $\gY$. Inspired by Eq.~\ref{eq:decomp}, we can estimate an optimal \(\tilde{\mathcal{X}}\) using decomposed shorter contexts \(\{\mathcal{X}_1, \ldots, \mathcal{X}_n\}\). Thus, \(I(\mathcal{X}; \tilde{\mathcal{X}})\) can be computed by processing each subset \(\mathcal{X}_i\) individually. However, as the number of subsets \(n\) increases, accounting for all preceding subsets becomes computationally demanding. To alleviate this burden, we propose reducing the number of conditional subsets considered by replacing the entire sequence of previous subsets with a compressed surrogate \(\hat{\mathcal{X}}_i\), which is iteratively derived using a compression function \(\hat{\mathcal{X}}_i = g(\hat{\mathcal{X}}_{i-1}, \mathcal{X}_{i-1})\). Consequently, Eq.~\ref{eq:decomp} can be reformulated as follows:
\begin{equation}
    I(\gX,\tilde{\gX}) = I(\gX_1,\cdots,\gX_n;\tilde{\gX}) \simeq I(\gX_1;\tilde{\gX})+ \sum\limits_{i=2}^n I(\gX_i; \tilde{\gX} | \hat{\gX}_i)).
    \label{eq:map}
\end{equation}
The equality can be upheld under two specific conditions: (1) the decomposed variables \(\{\mathcal{X}_1, \ldots, \mathcal{X}_n\}\) are mutually independent, and (2) the compression function \(g(\cdot)\) is optimally designed, ensuring that the compressed surrogate \(\hat{\mathcal{X}}_i\) encapsulates all relevant information from the preceding subsets with respect to \(\tilde{\mathcal{X}}\). Otherwise, $I(\gX,\tilde{\gX})$ can only be approximately estimated.

\paragraph{Empirical Analysis}
\begin{figure}
    \centering
    \includegraphics[width=0.8\linewidth]{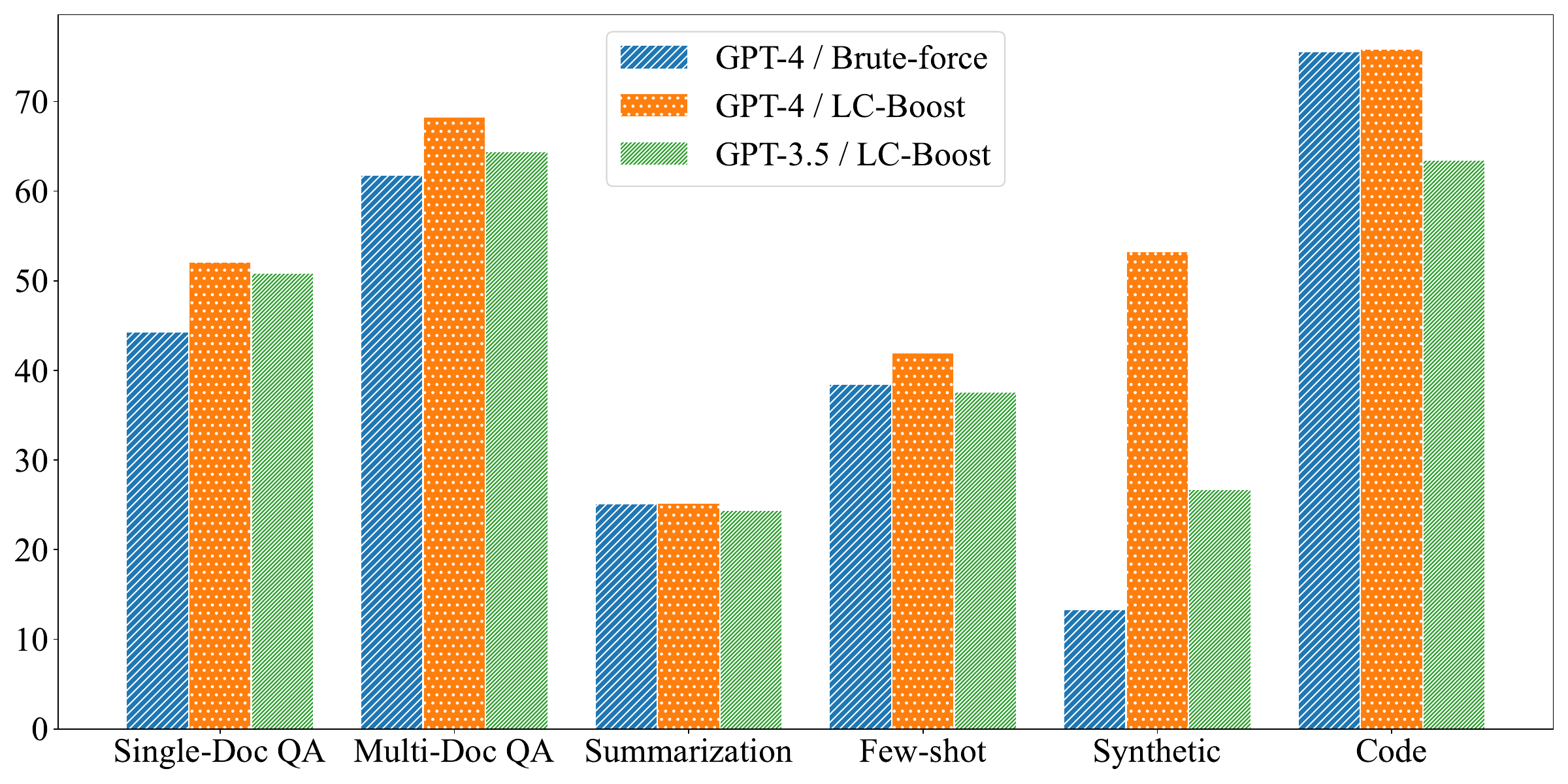}
    \caption{Pilot Study Across Various Tasks: In the Brute-force setting, the entire context is processed by GPT-4-128K. In the LC-Boost setting, the maximum context length is restricted to 4K, and LC-Boost is utilized to solve the long-context problem with short context. }
    \label{fig:explore}
    \vspace{-10pt}
\end{figure}

To empirically assess the accuracy of estimating the minimal necessary context \(\tilde{\mathcal{X}}\) using decomposed short contexts \(\{\mathcal{X}_1, \ldots, \mathcal{X}_n\}\), we conduct pilot experiments across various tasks requiring long contexts. Specifically, we utilize GPT-4-128K to perform these tasks in two settings: (1) feeding the entire long context into GPT-4-128K in a brute-force manner, instructing the model to directly produce the output answer, and (2) decomposing the full context into short contexts and applying the methods defined in Eq.~\ref{eq:map} to approximate \(\tilde{\mathcal{X}}\), which then guides the model to produce the final output (the LC-Boost setting).

Figure~\ref{fig:explore} presents the experiment results, which generally indicate that LC-Boost consistently performs as well as or better than the brute-force setting. In particular, for tasks such as QA, few-shot learning, and synthetic tasks, LC-Boost outperforms the brute-force setting. This is because the decomposed short contexts for these tasks are more likely to be mutually independent given the input query which can be adequately supported by a few extracted contexts from the long context. By precisely
locating these supported context, it can filter out irrelevant context of $\gX$ that might otherwise undermine task performance.
For tasks like summarization and code completion, the inherent properties of these tasks require considering the mutual dependencies among all decomposed short contexts, making the LC-Boost setting more challenging. However, as discussed in Eq.~\ref{eq:map}, when the compression function \(g(\cdot)\) is optimal, we can achieve the optimal \(\tilde{\mathcal{X}}\). GPT-4 serves as such a strong compression function, ensuring that the compressed surrogate \(\hat{\mathcal{X}}_i\) is well-estimated. Consequently, in these tasks, LC-Boost achieves performance that is equal to or slightly better than the brute-force setting.

Through theoretical analysis, we can posit that long-context tasks are short-context solvable if we can estimate a better minimal necessary context \(\tilde{\mathcal{X}}\) from the decomposed short contexts \(\{\mathcal{X}_1, \ldots, \mathcal{X}_n\}\) than from the long context \(\mathcal{X}\). Empirical analysis supports this assumption, demonstrating that in most cases, the estimation error of deriving \(\tilde{\mathcal{X}}\) from the long context \(\mathcal{X}\) is often larger than from the decomposed short contexts \(\{\mathcal{X}_1, \ldots, \mathcal{X}_n\}\). This indicates that using short contexts can be comparatively more advantageous than using the full context. Therefore, we can validate our argument in Section~\ref{sec:intro}: \textbf{most long-context tasks, if not all, are short-context solvable.}

\subsection{The Proposed Method: LC-Boost}
\label{sec:method}
We propose a method called LC-Boost, which utilizes short LLMs to solve general long-context tasks. LC-Boost begins with an input query \(q\) and a long context \(\mathcal{X}\), with the goal of producing an output answer \(\mathcal{Y}\). Since the underlying LLM in LC-Boost has a limited context size~(we limit LC-Boost working with 4K context length), directly generating the output answer \(\mathcal{Y}\) is infeasible for long-context tasks. To address this, we propose solving long-context tasks by strategically understanding the decomposed short contexts \(\mathcal{X} = \{\mathcal{X}_1, \cdots, \mathcal{X}_n\}\). From these short contexts, we aim to extract the minimal necessary context \(\tilde{\mathcal{X}}\) to support the generation of the output answer \(\mathcal{Y}\).

LC-Boost achieves this goal through a decision-making process involving iterative interactions between LC-Boost and the decomposed short contexts \(\{\mathcal{X}_1, \cdots, \mathcal{X}_n\}\) with respect to the input query \(q\). In the process, LC-Boost interact with each short context \(\mathcal{X}_i\), employing two types of actions: information access and information utilization. 

We denote an action at time step $i$ by $a_i$ and denote the relevant context LC-Boost obtains from the $i$-th short context $\gX_i$ by $\tilde{\gX}_i$
The action $a_i$ is predicted by considering the current short context $\gX_i$, the input query $q$, as well as all previous extracted relevant information $\tilde{\gX}_{1:i-1}$: $
    a_i=\gamma(q,\gX_i|\tilde{\gX}_{1:i-1}),
$
where $\gamma(\cdot)$ denotes LC-Boost's underlying LLM.

Predicting the action $a_i$ in a continuous space is challenging as it requires the underling model to reason about highly implicit relations among the input query, the current context, and the previous contexts. Therefore, we define a discrete action space $\gA$ comprising: 
(1)~\texttt{[Task Understanding]}: analyzing the query and task for initialization;
(2)~\texttt{[Retrieve]}: accessing text evidence by a retrieval method;
(3)~\texttt{[Move]}: accessing the next short text context directly;
These two are information access actions which define the LC-Boost's trajectory to access short contexts. 
(4)~\texttt{[Append]}: generating relevant context $\tilde{\gX}_i$ independently, denoting by $\tilde{\gX}_i=a_i(\gX_i)$;
(5)~\texttt{[Merge]}: generating relevant context $\tilde{\gX}_i$ with respect to previous extracted relevant information, denoting by $\tilde{\gX}_i=a_i(\gX_i|\tilde{\gX}_{1:i-1})$;
(6)~\texttt{[Answer]}: answering the user query and returning;
(7)~\texttt{[Aggregation]}: aggregating all relevant information and returning.
We define our LC-Boost frame in Algorithm~\ref{ago:framework}.

\begin{algorithm}[t]

\caption{LC-Boost Framework}
\small
\label{alg:LCBoost}
\begin{algorithmic}[1]
\STATE \textbf{Input:} Input query $q$, long context $\mathcal{X}$
\STATE \textbf{Output:} Answer $\mathcal{Y}$
\STATE Decompose $\text{long context}~\gX \leftarrow \{\gX_1,\cdots,\gX_n\}$
\STATE Initialize $\text{extracted relevant context}~\tilde{\gX}_0 \leftarrow \text{None}$
\STATE Perform \texttt{[Task Understanding]}
\WHILE{$i \leq n$}
    \STATE Select an action $a_i \leftarrow a_i=\gamma(q,\gX_i|\tilde{\gX}_{1:i-1}),~a_i\in \gA$
    \STATE{\textbf{if} $a_i$ is \texttt{[Move]}} \textbf{then} $i\leftarrow i+1$, continue
    \STATE{\textbf{if} $a_i$ is \texttt{[Retrieve]}} \textbf{then} retrieve evidence from $\gX=\{\gX_1,\cdots,\gX_n\}$
    \STATE{\textbf{if} $a_i$ is \texttt{[Append]}} \textbf{then} generate relevant context by $\tilde{\gX}_i=a_i(\gX_i)$
    \STATE{\textbf{if} $a_i$ is \texttt{[Merge]}} \textbf{then} generate relevant context by $\tilde{\gX}_i=a_i(\gX_i|\tilde{\gX}_{1:i-1})$
    \STATE{\textbf{if} $a_i$ $\in\{$\texttt{[Answer]}$,$\texttt{[Aggregation]}$\}$} \textbf{then} generate answer $\gY=\gamma(q,\tilde{\gX}_{1:i})$, \textbf{break}
    \STATE $i\leftarrow i+1$
\ENDWHILE
\STATE \textbf{return} answer $\gY$
\end{algorithmic}

\label{ago:framework}

\end{algorithm}

Though the pre-defined action space \(\mathcal{A}\) comprises only seven actions, LC-Boost serves as a general framework sufficient for solving most long-context tasks. This effectiveness is based on the following reasons: \textbf{(1)~Flexible accessibility:}~By utilizing both \texttt{[Retrieve]} and \texttt{[Move]} actions, LC-Boost can access any short context \(\mathcal{X}_i \in \mathcal{X}\) in a flexible trajectory, avoiding the need to browse the entire long context. This makes the information accessing process more efficient. \textbf{(2)~Accurate information acquisition:}~Through the \texttt{[Append]} and \texttt{[Merge]} actions, LC-Boost can either independently extract relevant information from the current short context, appending it to previously extracted information, or merge the current relevant information into the previous relevant information. This capability allows LC-Boost to acquire relevant information in a compatible manner, making it adaptable to many knowledge-intensive tasks. and \textbf{(3)~Dynamic answering:}~Using the \texttt{[Answer]} and \texttt{[Aggregate]} actions, LC-Boost can dynamically utilize the acquired relevant information to produce the target form of the answer (e.g., a short answer for QA tasks via the \texttt{[Answer]} action, or a long answer for summarization tasks via the \texttt{[Aggregate]} action). 

In our pilot study depicted in Figure~\ref{fig:explore}, we observe that while GPT-3.5 serves as an inferior foundation model compared to GPT-4, it still demonstrates significant effectiveness when incorporated with LC-Boost. Given considerations of efficiency and cost-effectiveness, we employ GPT-3.5 as the foundation model for LC-Boost in the subsequent experiments. Besides, we show the prompts used in LC-Boost in Appendix~\ref{sec:prompt}.

\section{Experiments}

\subsection{Experiment Settings}
We evaluate LC-Boost and baseline models on 12 datasets, including:
(1) Single-Doc QA: NarrativeQA~\citep{kočiský2017narrativeqa}, Qasper~\citep{dasigi2021dataset}, and MultiFieldQA~\citep{bai2023longbench}.
(2) Multi-Doc QA: HotpotQA~\citep{yang2018hotpotqa}, 2WikiMQA~\citep{ho-etal-2020-constructing}, and MuSiQue~\citep{trivedi2022musique}.
(3) Summarization: GovReport~\citep{huang-etal-2021-efficient} and MultiNews~\citep{fabbri2019multinews}.
(4) Few-shot Learning: SAMSum~\citep{gliwa-etal-2019-samsum}.
(5) Synthetic Task: Passage Count~\citep{bai2023longbench} and Self-Constructed Dataset.
(6) Code Completion: LCC~\citep{guo2023longcoder}.
 More details about the evaluation datasets and metrics are introduced in Appendix~\ref{sec:data_detail}.

We compare our LC-Boost with three types of models: (1) Short LLMs (defined as with context length $<$ 32K): Llama2-7B-Chat-4K~\citep{llama}, Llama3-8B-Instruct-8K and Vicuna-v1.5-7B-16K~\citep{vicuna2023}; (2) Long LLMs (defined as with context length $\ge$ 32K): LongChat-v1.5-7B-32K~\citep{longchat2023}, Mistral-7B-Instruct-v0.2-32K~\citep{jiang2023mistral}, Llama3-8B-80K~\cite{zhang2024extending}, Phi-3-mini-128K~\citep{abdin2024phi3} and Yi-9B-200K~\citep{ai2024yi}; (3)~Closed-Source LLMs: DeepSeek-v2~(236B MoE model, ranks top-tier in MT-Bench)~\citep{deepseekai2024deepseekv2}, Claude-3-Haiku\footnote{\url{https://www.anthropic.com/claude}} and GPT-3.5-turbo-16K\footnote{\url{https://platform.openai.com/docs/models}}. In the experiments, if the context length exceed the model's length limit, following~\citet{bai2023longbench}, we truncate the context from the middle since the front and end of the context may contain crucial information. We provide further implementation details in Appendix~\ref{sec:prompt}.

\subsection{Main Results}
\begin{table*}[t]
    \centering
    \small
    \caption{Main experiment results. The best results are in bold and the secondary results are marked with underline. We  report the average scores (\%) on the main tasks. The detailed scores over all dataset are shown in Table~\ref{tab:exp_detail}.}
    
\begin{tabular}{lccccccc}

\toprule
Models  & Single-Doc & Multi-Doc & Summ. & Few-shot & Synthetic & Code  \\
\midrule
\multicolumn{3}{l}{\textbf{Short LLMs (Context Length $<$ 32K)}} & \\
\midrule
Llama2-7B-Chat-4K & 24.9 & 22.5 & 26.6 & 40.7 & 6.3& 52.4\\
Llama3-8B-Instruct-8K &  37.3 & 36.0 & 26.5 & 42.7 & 15.0 & 57.5\\ 
Vicuna-v1.5-7B-16K & 28.0 & 18.6 & 27.5 & 40.8 & 8.9 & 51.0\\

\midrule
\multicolumn{3}{l}{\textbf{Long LLMs (Context Length $\ge$ 32K)}} & \\
\midrule
LongChat-v1.5-7B-32K & 28.7 & 20.6 & 28.6 & 34.2 & 6.8 & 53.0\\ 
Mistral-7B-Instruct-v0.2-32K & 31.9 & 26.0 & 29.3 & \underline{43.0} & 14.0 & 55.4 \\
Llama3-8B-80K & 43.6 & 43.1 & 30.2 & 42.9 & 19.6 & 53.6 \\
Phi-3-mini-128K & 33.5 & 38.2 & 28.8 & 36.0 & 19.9 & \underline{60.1}\\
Yi-9B-200K & 29.6 & 38.7 & 28.4 & 14.6 & 6.5 & \textbf{72.1}\\
\midrule
\multicolumn{3}{l}{\textbf{Closed-Source LLMs}} & \\
\midrule

DeepSeek-v2 (32K) & 37.6 & \underline{49.1} & \underline{30.8} & 39.3 & 14.5 & 37.0  \\

Claude-3-Haiku (200K) &  \underline{41.9} & 45.4 & 30.1 & 7.2 & \underline{25.5} & 16.9 \\
GPT-3.5-turbo-16K &  39.8& 38.7 & 28.1 & 41.7 & 18.7 & 54.7 \\

\midrule

LC-Boost~(4K)  & \textbf{47.8} & \textbf{56.4} &\textbf{31.8} & \textbf{44.1} & \textbf{27.5} & 59.0 \\

\bottomrule
\end{tabular}
\label{tab:exp}
\end{table*}

Table~\ref{tab:exp} shows the overall experimental results for all models across all tasks. From the table, we derive several key findings:
\textbf{First}, LC-Boost, with a context length of 4K, outperforms all baseline models in all tasks except for the Code Completion task. This result verifies LC-Boost's capability to effectively solve long-context tasks by strategically processing decomposed short contexts.
\textbf{Second}, long LLMs generally perform better than short LLMs, indicating the effectiveness of fine-tuning LLMs to adapt to long contexts. However, the performance of long LLMs is not consistently stable across different tasks. For example, Yi-9B-200K excels in the Code Completion task but does not show consistent performance in other tasks such as single-doc QA, few-shot learning, and synthetic tasks. This inconsistency suggests that adapting LLMs to long contexts may compromise their general abilities.
\textbf{Last}, LC-Boost consistently surpasses its underlying LLM, GPT-3.5-turbo-16K, across all tasks by a notable margin. This demonstrates that LC-Boost can achieve improved performance while simultaneously reducing resource costs, making LC-Boost an environmentally friendly method.

\subsection{Ablation Study: Dynamic is Important}
To investigate the necessity of LC-Boost's design, we conduct ablation studies by changing LC-Boost's action space \(\mathcal{A}\), resulting in different information acquisition strategies. We experiment with the following settings: (1)~\texttt{[Retrieve]} only: Directly retrieve the most relevant short context. (2)~\texttt{[Merge]} only: Sequentially process all short contexts while considering the previously processed context. (3)~\texttt{[Append]} only: Sequentially process all short contexts independently. (4)~\texttt{[Merge]} \& \texttt{[Move]}: Selectively process short contexts while considering the already processed context. (6)~\texttt{[Append]} \& \texttt{[Move]}: Selectively process short contexts independently. (7): \texttt{[Retrieve]} \& \texttt{[Move]}: Retrieve the top-\(k\) relevant short contexts and selectively process a few of them. (8): Brute-force: Directly produce the answer based on the entire long context. (9)~Random: For each short context, randomly select an action. Based on the acquired information from each strategy, LC-Boost then selects either the \texttt{[Answer]} or \texttt{[Aggregation]} action to produce the final answer.

\begin{figure}[t]
    \centering
    \includegraphics[width=0.9\linewidth]{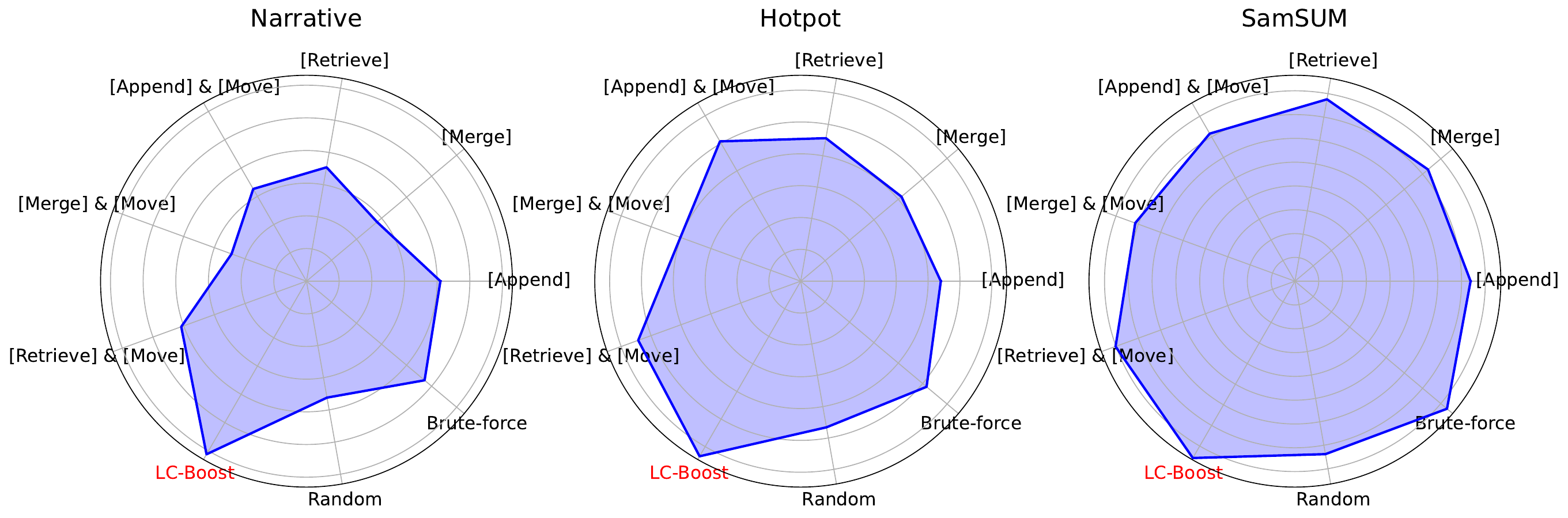}
    \caption{Performance comparison on different context processing strategies in the ablation study. NarrativeQA~(left) is a single-doc QA task. HotpotQA~(middle) is a multi-doc QA task. SamSUM~(right) is a few-shot learning task.}
    \label{fig:abl}
\end{figure}

Figure~\ref{fig:abl} illustrates the results, from which we find that: (1)~Compared to fixed processing strategies, LC-Boost customizes the action trajectory for each query, resulting in notable performance improvements. This finding emphasizes the importance of the dynamic capabilities of LC-Boost. (2)~LC-Boost is particularly effective in single-doc QA and multi-doc QA tasks, as it can accurately select the minimal necessary context required to answer the input query, filtering out irrelevant information from the long context. (3)~In the few-shot learning task, LC-Boost does not significantly outperform the fixed strategies. This is attributed to the numerous in-context examples provided within the task, which offer substantial guidance, thus diminishing the impact of the number of in-context examples on the final performance.

\subsection{Case Study: Model Behavior Analysis on Self-Construct Dataset}

\begin{table*}[t]
\small
    \centering
    \caption{Case study on the self-constructed dataset. Correct answers are marked in \textcolor{teal}{teal}, incorrect answers in \textcolor{red}{red}, and ambiguous answers in \textcolor{orange}{orange}.}
    \begin{tabular}{p{.98\linewidth}}
    \toprule
     \textbf{Query}: How many papers in ACL 2023 only have one author? \\
     \textbf{Context}: Full accepted paper list in ACL 2023 main conference. (Context length: 45K)\\
     \textbf{Ground-truth target}:  8 papers\\
\midrule
   \textbf{Phi-3-mini-128K}: \textcolor{red}{11 papers}  \textbf{GPT-3.5-turbo-16K}: \textcolor{red}{0 papers}  \textbf{Claude-3-Haiku-200K}: \textcolor{red}{1 papers}  (Acc. Score: 0)\\
\midrule
    \textbf{LC-Boost's action trajectory}:  \texttt{[Task Reasoning]} $\rightarrow$ \texttt{[Append]}$\rightarrow\cdots$ $\rightarrow$ \texttt{[Append]}$\rightarrow$ \texttt{[Aggregation]}\\
     \textbf{LC-Boost}: \textcolor{teal}{8 papers} (Acc. Score: 1)\\

     \midrule
     \midrule
     \textbf{Query}: List all people names that are petrified, separated by comma. \\
     \textbf{Context}: Full content of Harry Potter and the Chamber of Secrets. (Context length: 122.6K)\\
     \textbf{Ground-truth target}: Colin Creevey, Justin Finch-Fletchley, Penelope Clearwater, Hermione Granger\\
\midrule
   \textbf{Phi-3-mini-128K}: \textcolor{teal}{Hermione Granger}, \textcolor{red}{Ginny Weasley}, \textcolor{orange}{Mrs Norris} (F1-Score: 0.29)\\
   \textbf{GPT-3.5-turbo-16K}: \textcolor{teal}{Colin Creevey}, \textcolor{orange}{Mrs Norris} (F1-Score: 0.33)\\
   \textbf{Claude-3-Haiku-200K}: \textcolor{orange}{Nick}, \textcolor{teal}{Hermione}, \textcolor{red}{Ron} (F1-Score: 0.18)\\
\midrule
    \textbf{LC-Boost's action trajectory}:  \texttt{[Task Reasoning]} $\rightarrow$ \texttt{[Move]}$\rightarrow\cdots$ $\rightarrow$ \texttt{[Merge]}$\rightarrow$ \texttt{[Aggregation]}\\
     \textbf{LC-Boost}: \textcolor{teal}{Colin Creevey}, \textcolor{teal}{Penelope Clearwater}, \textcolor{teal}{Hermione Granger}, \textcolor{orange}{Nick}, \textcolor{orange}{Mrs Norris} (F1-Score: 0.71)\\
    \bottomrule
    \end{tabular}
    \label{tab:case}
\end{table*}
In Table~\ref{tab:case}, we present two case studies from the self-constructed dataset. These cases are particularly challenging as they require reasoning across the entire long context. Despite having sufficient context size, LLMs struggle to generate correct responses. In contrast, LC-Boost dynamically customizes solutions for each case, thereby effectively solving the problems using a shorter context length.

For the first query,  LC-Boost performs \texttt{[Append]} or \texttt{[Move]} actions across all short context along with a rewritten query, "Extract paper information in the following list that have only one author," derived via \texttt{[Task Reasoning]}. After processing all short contexts, LC-Boost employs the \texttt{[Aggregation]} action to compile the final answer. This approach simplifies the task compared to directly extracting a numeric answer from the entire long context, mimicking the human process of reading comprehension and thereby producing accurate results.

In the second case, the query necessitates conditional reasoning on each short context. As highlighted in previous research \citep{liu2023lost}, reasoning directly from the entire context risks losing crucial information, particularly in the middle of the long context. Thus LLMs tend to miss key details such as people's names. LC-Boost addresses this issue by processing only one short context at a step where it extracts information from arbitrary position of the long text with equal accuracy. 
Additionally, answers marked in \textcolor{orange}{orange} include non-human names (\eg, cat, ghost) that are misconstrued as people, illustrating a common challenge where models fail to differentiate in-depth entity properties.

\subsection{Context be Short, Energy be Saved!}
\label{sec:energy}
\begin{wrapfigure}[14]{r}{0.45\textwidth}
\centering
\vspace{-5pt}
\includegraphics[width=\linewidth]{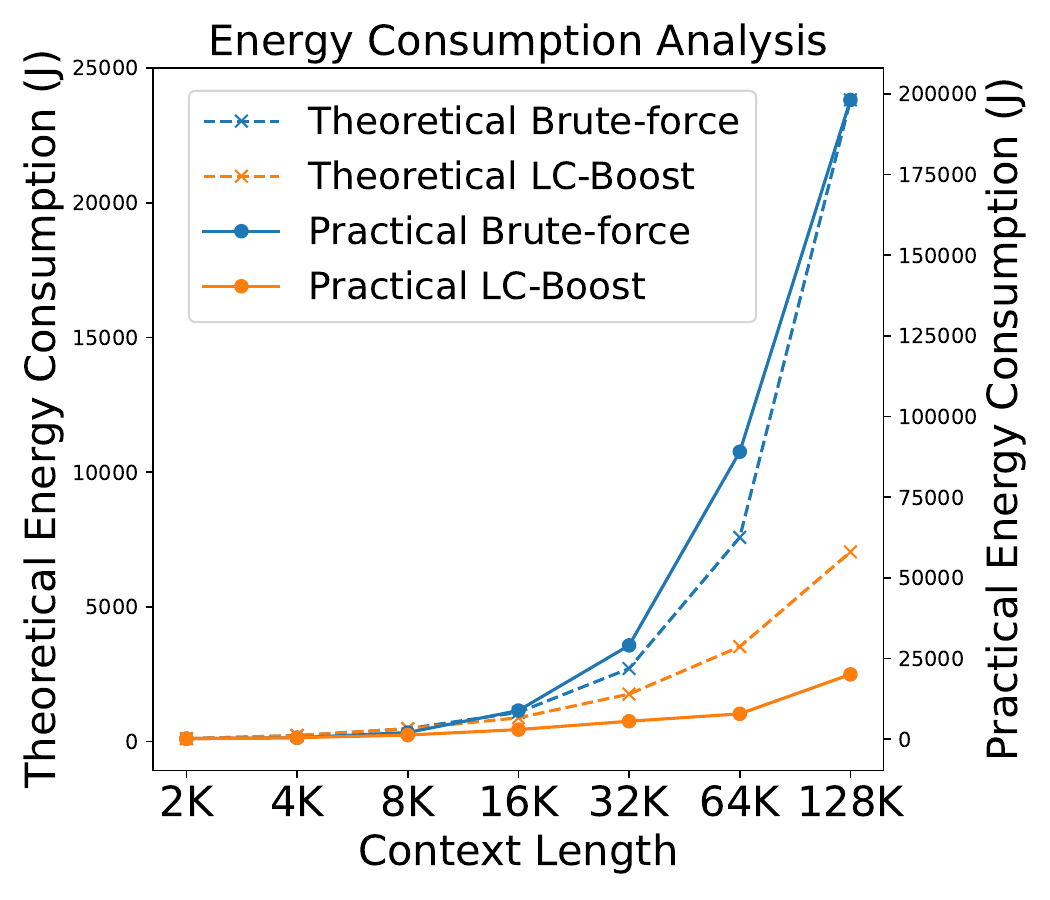}
\caption{Energy consumption analysis.\label{fig:resource}}
\vspace{-5pt}
\end{wrapfigure}

Recently, we have witnessed the remarkable success of LLMs, which are becoming an indispensable part of our daily lives. We believe that in the near future, LLMs will become as ubiquitous as electricity or gas supply, serving as fundamental infrastructure in human society. At that point, the energy consumption of LLMs will emerge as a significant environmental concern. Therefore, it is imperative for the research community to focus on reducing the energy consumption associated with these models.
Figure~\ref{fig:resource} presents an analysis of energy consumption, comparing the brute-force approach with our LC-Boost method. The $y$-axis is measured in Joules. The theoretical energy consumption is estimated for 7B LLMs across varying context lengths. We roughly estimate the energy consumption using the formula \(\left(\frac{\text{Total Float Operation}}{312 \text{ TFLOPS}}\right) \times 400W\), assuming the use of an A100 GPU with a compute capability of 312 TFLOPS for BFLOAT16 operations and a maximum TDP of 400W\footnote{The calculation of total float operations is based on the method outlined in \url{https://www.harmdevries.com/post/context-length/}}. The practical energy consumption is estimated by recording the GPU time and GPU power during inference with different context lengths. We use a Llama2-7B-128K~\citep{peng2023yarn} and a Llama2-7B-chat-4K~\citep{touvron2023llama} for the brute-force setting and LC-Boost, respectively.
Figure~\ref{fig:resource} clearly indicates that longer context lengths significantly increase energy consumption with the brute-force method, especially evident in practical measurements. This difference is primarily due to the need to distribute sequence activation tensors across multiple GPUs in practical experiment, with tensor I/O exacerbating inference latency and thereby inflating energy costs. In contrast, our LC-Boost method, working with 4K context lengths, shows only a mild increase in energy consumption across contexts, thereby confirming its energy efficiency while maintaining comparable or superior performance on long-context tasks. We also provide an analysis on token consumption in Appendix~\ref{sec:token}.

\section{Related Works}

Dealing with long contexts is a fundamental research problem for LLMs, as many real-world applications involve long-context inputs \citep{li2023long,fu2024data}. The most direct approach to address long-context tasks is to increase the working context size of LLMs \citep{abdin2024phi3,ai2024yi,li2023long,cai2024internlm2}. A year ago, significant research efforts focused on extending the working context size of LLMs from 4K to 32K \citep{jiang2023mistral,longchat2023,chen2023longlora,du2022glm}. Currently, many popular open-source and close-source LLMs still operate with a context size under 32K \citep{touvron2023llama,gpt-4}, such as GPT-3.5-turbo, which has a 16K context length. Recently, research has shifted towards extending LLMs' working context to the million-level. Notably, GPT-4 was updated to a 128K context length not long ago, and the newly released GPT-4o also operates with a 128K context. Moreover, several recent open-source LLMs have been introduced with context lengths exceeding 100K, for example, the Yi series model supports up to 200K \citep{ai2024yi}, and the Phi-3 model operates with 128K \citep{abdin2024phi3}.

Instead of merely increasing the context length, another approach to address long-context tasks involves extracting a short surrogate context from the full context. This includes techniques like retrieval-augmented generation (RAG) and context refinement methods \citep{izacard2021leveraging,gao2024retrievalaugmented,wang2023learning,qian2024grounding}. However, many of these methods utilize task-specific strategies to manage the long context. For instance, RAG methods often deploy retrievers to select relevant context chunks as supporting evidence \citep{izacard2021distilling,xu2023retrieval,jiang2023active}. Recent studies have criticized the chunking process in RAG for undermining the semantic coherence of the long context and have proposed chunking-free methods to refine the long context into a concise surrogate context \citep{qian2024grounding, luo2024bge}.
Furthermore, some studies have also explored sequential processing strategies, such as \citet{pcw} and \citet{longctx}, to sequentially process the context in a manner that preserves its integrity. 

Lastly, reasoning-based methods also show significant potential for addressing long context tasks \citep{nakano2022webgpt,yang2023auto,driess2023palm}. These methods predominantly employ a decision-making process to navigate through the long context sequentially, utilizing reasoning techniques such as in-context learning \citep{dong2022survey}, chain-of-thought \citep{wei2022chain}, and self-reflection \citep{shinn2023reflexion}. In this paper, LC-Boost incorporates a decision-making process that dynamically customizes the action trajectory for each query, thereby offering considerable flexibility in accessing and leveraging information to produce the final output answer.

\vspace{-5pt}
\section{Conclusion}
In this paper, we argue that most long-context tasks are short-context solvable, and we validate this claim through both theoretical and empirical analysis. We propose a method called LC-Boost to solve long-context tasks by decomposing the long context into short contexts and processing them using a decision-making process. We conduct experiments on 12 datasets to compare LC-Boost with long LLMs and other baseline models. Empirical results verify LC-Boost’s effectiveness in solving long-context tasks. Additionally, we discuss the energy consumption of LC-Boost versus long LLMs, demonstrating that LC-Boost can achieve comparable performance with significantly less energy consumption. In Appendix~\ref{sec:limit}, we also discuss the limitations and broader impact of this paper.
\clearpage 
\bibliography{main}
\bibliographystyle{unsrtnat}
\clearpage
\appendix

\begin{table*}[t]
\label{sec:exp_detail}
\caption{Main experiment results. The best results are in bold and the secondary results are marked
with underline. We report the average scores (\%) on all tasks.}
\small
    \centering
    \begin{tabular}{lccccccc}
    \toprule
         Model   & Narrative & Qasper & MultiField & Hotpot & MuSiQue & 2Wiki \\
    \midrule
    \multicolumn{3}{l}{\textbf{Short LLMs (Context Length $<$ 32K)}} & \\
    \midrule
         Llama2-7B-Chat-4K & 18.7&19.2&36.8&25.4&9.4&32.8\\
Llama3-8B-Instruct-8K &  21.5 & 43.0 & 47.5 & 47.3 & 23.3 & 37.5\\ 
Vicuna-v1.5-7B-16K & 19.4&26.1&38.5&25.3&9.8&20.8\\

\midrule
\multicolumn{4}{l}{\textbf{Long LLMs (Context Length $\ge$ 32K)}} & \\
\midrule

LongChat-v1.5-7B-32K & 16.9&27.7&41.4&31.5&9.7&20.6\\ 
Mistral-7B-Instruct-v0.2-32K &  21.6 & 29.2 & 47.9 & 37.7 & 18.6 & 21.8\\
Llama3-8B-80K &  28.8 & 47.4 & \underline{54.5} & 55.8 & 27.4 & 46.0\\
Phi-3-mini-128K & 21.0 & 39.4 & 51.5 & 48.1 & 28.2 & 38.1\\
Yi-9B-200K & 15.6 & 39.3 & 33.8 & 51.4 & 26.6 & 38.2\\
\midrule
\multicolumn{3}{l}{\textbf{Closed-Source LLMs}} & \\
\midrule

DeepSeek-v2 (32K) &  18.3 & \underline{45.7} & 48.9 & \underline{57.7} & 22.6 & \textbf{66.9}\\

Claude-3-Haiku (200K) &  \underline{30.2} & 44.0 & 51.5 & 51.5 & \underline{32.5} & 52.1 \\
GPT-3.5-turbo-16K &  23.6&43.3&52.3&51.6&26.9&37.7\\
\midrule
LC-Boost (4K) &\textbf{30.6} & \textbf{50.6} & \textbf{62.1}& \textbf{63.5} & \textbf{42.5} & \underline{63.1}\\
\midrule
\midrule
Model   & GovReport&MultiNews & SAMSum &  LCC &PCount& Self\\
\midrule
    \multicolumn{3}{l}{\textbf{Short LLMs (Context Length $<$ 32K)}} & \\
    \midrule
         Llama2-7B-Chat-4K & 27.3&25.8&40.7&52.4&2.1&10.5\\
Llama3-8B-Instruct-8K &30.1 & 27.6 & 42.7 & 57.5 & 8.0  &21.9 \\ 
Vicuna-v1.5-7B-16K & 27.9&27.2&40.8&51.0&6.5&11.3\\

\midrule
\multicolumn{4}{l}{\textbf{Long LLMs (Context Length $\ge$ 32K)}} & \\
\midrule

LongChat-v1.5-7B-32K & 30.8&26.4&34.2&53.0&1.0 & 12.5\\ 
Mistral-7B-Instruct-v0.2-32K &  31.7 & 26.9 & 43.0 & 55.4 & 2.6 &25.4\\
Llama3-8B-80K &  32.3 & \underline{28.1} & \underline{42.9} & 53.6 & 3.5 & 35.7\\
Phi-3-mini-128K & 32.6 & 24.9 & 36.0 & \underline{60.1} & 3.2&36.5\\
Yi-9B-200K & 30.3 & 26.5 & 14.6 & \textbf{72.0} & 4.2&8.7\\
\midrule
\multicolumn{3}{l}{\textbf{Closed-Source LLMs}} & \\
\midrule

DeepSeek-v2 (32K) &  \textbf{35.2} & 26.3 & 39.3 & 37.0 & \textbf{12.7}&16.2\\

Claude-3-Haiku (200K) &  34.1 & 26.1 & 7.2 & 16.9 & 5.0 & \underline{46.0}\\
GPT-3.5-turbo-16K &  29.5&26.7&41.7&54.7&4.5&32.9\\
\midrule
LC-Boost (4K) &\underline{34.4} & \textbf{29.2}& \textbf{44.1 }& 59.0 & \underline{7.2}& \textbf{47.7}\\
    \bottomrule
    \end{tabular}
    
    \label{tab:exp_detail}
\end{table*}

\section{More details of the Datasets}
\label{sec:data_detail}

\begin{table*}[t]
\caption{Statistical information of the datasets utilized in this paper.}
\small
    \centering
    \begin{tabular}{cccccccc}
    \toprule
         Dataset   & Narrative & Qasper & MultiField & Hotpot & MuSiQue & 2Wiki \\
    \midrule
         Num of Samples & 200 & 200 & 150 & 200 & 200 & 200 \\
         Ave. Length &  18,409 & 3,619 & 4,559 & 9,151 & 11,214 & 4,887  \\
         Metric & F1 & F1 & F1 & F1 & F1 & F1\\
\midrule
\midrule
Dataset   & GovReport&MultiNews & SAMSum & PCount & LCC & Self\\
\midrule
Num of Samples& 200 & 200 & 200 & 200 & 500  & 32\\
Ave. Length & 8,734& 2,113 & 6,258 & 11,141 & 1,235 & 39,420\\
Metric & Rouge-L & Rouge-L &Rouge-L & Accuracy & Edit Sim & F1\&Accuracy\\
    \bottomrule
    \end{tabular}
    
    \label{tab:data}
\end{table*}
We evaluated all models on 12 datasets, as shown in Table~\ref{tab:data}. Most of these datasets are provided by the LongBench benchmark~\citep{bai2023longbench}. Following LongBench, we used F1-score, accuracy, and edit similarity as the evaluation metrics. Additionally, we manually annotated a self-constructed dataset comprising long contexts from practical scenarios, such as the full schedule of the Olympic Games and the complete list of accepted papers at ACL. The queries in the self-constructed dataset involve reasoning over the entire long context. For example, “Who has the most accepted papers at ACL 2023?” These queries require the model to accurately understand the long context and perform reasoning, making them highly challenging. The details of the self-constructed dataset are in Table~\ref{tab:self}.

\begin{table*}[t]
\caption{Data details of the self-constructed dataset.}
\small
    \centering
    \begin{tabular}{p{3.5cm}p{1.5cm}p{1.2cm}p{5cm}}
    \toprule
        Source & Length & \# Queries & Example Query \\
    \midrule
        Accepted paper list of ACL 2023 Main Conference & 44,490 & 7 & Who has the most accepted paper in ACL 2023? \\
        \midrule
        The Diamond Sutra & 19,993 & 3 & How many chapters of the Sutra? \\
        \midrule
        Schedule of The 2024 Olympic Games & 15,844 & 9 & Which day has the most gold medal events? \\
        \midrule
        Subtitle of The Big Bang Theory S3E14 & 11,136 & 6 & How long does this episode? \\
        \midrule
        The Little Prince & 22,471 & 4 & How many planets does the little prince visit? \\
        \midrule
        Harry Potter and the Chamber of Secrets & 122,591 & 3 & How many times has the chamber of secret been opened? \\
    \bottomrule
    \end{tabular}
    
    \label{tab:self}
\end{table*}

\section{Implementation Details}
\label{sec:prompt}

LC-Boost begins with the \texttt{[Task Understanding]} action after receiving the input query and context, using the prompt shown in Table~\ref{tab:task_understanding}. If the task does not include an input query, the two columns "Below is the query" and "\{input\_query\}" are omitted. Besides, for the synthetic task, we use the prompt shown in Table~\ref{tab:query_understanding} to reformulate the query for better adaptation to LC-Boost. Based on the output of the \texttt{[Task Understanding]} action, LC-Boost adopts different strategies to perform the task. Specifically, ``option [1]'' directs LC-Boost to utilize a retriever to rank all chunks of the long context. In this paper, we employ BGE-Reranker-Large as the retriever~\cite{bge_m3}. For ``option [2]'' and ``option [3]'', LC-Boost uses the prompts shown in Table~\ref{tab:seq_sum} and Table~\ref{tab:extract} to sequentially process each short context, respectively. After processing each short context, if the output is not "null", the newly summarized context is added to the "previous summarization". 

Once all short contexts are processed, LC-Boost aggregates all relevant information to produce the final answer. At this stage, we use the prompt provided by LongBench, replacing the full context with the surrogate context produced by LC-Boost. For ``option [4]'', LC-Boost utilizes the prompts provided by LongBench to process each short context and produces the answer as soon as the proper information is found. Table~\ref{tab:mul} presents an example prompt from LongBench, designed for MultiFieldQA tasks. We modified the prompt by adding the instruction ``If no answer can be found in the text, please output "null"''. This allows LC-Boost to skip irrelevant short contexts, performing the \texttt{[Move]} action. Specifically, for the Code Completion task, LC-Boost reversely browses the context code from near to far as the near context are more useful to predict the code completion. 
We evaluate all baseline models following the settings provided in LongBench~\footnote{\url{https://github.com/THUDM/LongBench}}. We use a node with 8 A100 80G GPUs to conduct all experiments.

\section{Token Consumption Analysis} \label{sec:token}
In Section~\ref{sec:energy}, our analysis confirms that LC-Boost significantly reduces energy consumption compared to long LLMs. However, most closed-source LLMs, such as the underlying model of LC-Boost, GPT-3.5-turbo, charge based on token consumption, \eg, US\$0.50 per 1M tokens for input and US\$1.50 per 1M tokens for output\footnote{\url{https://openai.com/api/pricing/}}. Consequently, it is crucial to examine whether the decision-making process of LC-Boost increases token consumption compared to the brute-force method.

To address this issue, we recorded the end-to-end token consumption for three datasets: NarrativeQA, GovReport, and LCC. After token counting, we conclude that LC-Boost’s token consumption was 34.1\% of the brute-force method’s consumption in NarrativeQA, 112\% in GovReport, and 29.5\% in LCC. These results indicate that LC-Boost’s token consumption varies significantly across different tasks. For tasks requiring precise context location, such as QA and code completion, LC-Boost can respond as soon as the relevant context is identified, thereby avoiding the need to process the full context. However, for tasks that necessitate information aggregation, such as summarization, LC-Boost may require more tokens for prompts in each iteration.
In practice, for token-consumption-sensitive LLMs, there might be a trade-off between performance and cost-efficiency, which also varies considerably across different tasks.
\begin{table}
\centering
\small
\caption{Prompt Template for the \texttt{[Task Understanding]} action.\label{tab:task_understanding}}
\begin{tabular}{l}
\toprule  
\makecell[l]{\textit{You need to process a task with a long context that greatly exceeds your context limit.} \\ 
\textit{The only feasible way to handle this is by processing the long context chunk by chunk.}} \\  

\makecell[l]{Below is the original task prompt: \\ \{task\_prompt\}} \\

\makecell[l]{Below is the query: \\ 
\{input\_query\}} \\

\makecell[l]{You have the following options to process the long context. Choose one of them:} \\
\makecell[l]{[1]. Retrieve the chunk most relevant to the input query to support answer generation.} \\
\makecell[l]{[2]. Summarize each chunk and then aggregate the summaries after processing all chunks.} \\
\makecell[l]{[3]. Extract key sentences from each chunk and then aggregate them after processing all chunks.} \\
\makecell[l]{[4]. Sequentially scan chunks and produce the answer as soon as the query can be answered.} \\
\makecell[l]{Below are some examples for reference: } \\ \makecell[l]{The examples begin as follows: \\ \{examples\} \\ The examples conclude here.} \\
\makecell[l]{Please learn the examples and select one of the options by only outputting the corresponding index number.} \\
\bottomrule
\end{tabular}
\end{table}

\begin{table}
\centering
\small
\caption{Query Rewritten Prompt Template for the \texttt{[Task Understanding]} action.\label{tab:query_understanding}}
\begin{tabular}{l}
\toprule  
\makecell[l]{\textit{You need to process a task with a long context that greatly exceeds your context limit.} \\ 
\textit{The only feasible way to handle this is by processing the long context chunk by chunk.}} \\  

\makecell[l]{Below is the original task prompt: \\ \{task\_prompt\}} \\

\makecell[l]{Below is the query: \\ 
\{input\_query\}} \\

\makecell[l]{You will process the long context with the following strategy:\\ \{strategy\}} \\
\makecell[l]{Do you think the the query is proper for processing context chunk? If not, rewrite the query.} \\
\makecell[l]{Below are some examples for reference: } \\ \makecell[l]{The examples begin as follows: \\ \{examples\} \\ The examples conclude here.} \\
\makecell[l]{Please study the examples carefully. If the query needs to be rewritten, directly output the revised query.\\ If no revision is necessary, output “null”.} \\
\bottomrule
\end{tabular}
\end{table}

\begin{table}
\centering
\small
\caption{Prompt Template for the \texttt{[Append]} action.\label{tab:extract}}
\begin{tabular}{l}
\toprule  
\makecell[l]{\textit{You are given an article and a question. Read the article carefully and follow my instructions to process it.}} \\  

\makecell[l]{Article:} \\
\makecell[l]{The article begins as follows: \\ \{article\} \\ The article concludes here.} \\

\makecell[l]{Question:} \\
\makecell[l]{\{question\}} \\

\makecell[l]{Instructions:} \\
\makecell[l]{Each sentence in the article is marked with a sentence identifier [si], for example [s1].} \\
\makecell[l]{Select up to ten key sentences from the article that are most likely to answer the question.} \\
\makecell[l]{Only output the selected sentence identifiers, separated by commas.} \\
\makecell[l]{Example: [s39],[s54]} \\
\makecell[l]{If no sentences are relevant, please output "null".} \\
\bottomrule
\end{tabular}
\end{table}
\begin{table}[t]
\centering
\small
\caption{Prompt Template for the MultiFieldQA Task from the LongBench Benchmark. Additions made by us are highlighted in blue.\label{tab:mul}}
\begin{tabular}{l}
\toprule  
\makecell[l]{\textit{
Read the following text and answer briefly.}} \\  

\makecell[l]{\{context\}} \\
\makecell[l]{Now, answer the following question based on the above text, only give me the answer and do not output any } \\
\makecell[l]{other words. \textcolor{blue}{ If no answer can be found in the text,  please output "null".}} \\
\makecell[l]{Question:\{question\}} \\
\makecell[l]{Answer:} \\

\bottomrule
\end{tabular}
\vspace{-10pt}
\end{table}

\begin{table}[t!]
\centering
\small
\caption{Prompt Template for the \texttt{[Merge]} action.\label{tab:seq_sum}}
\begin{tabular}{l}
\toprule  
\makecell[l]{\textit{You are provided with a portion of an article, a question, and summarization of the article's previous portions. }} \\  
\makecell[l]{\textit{Read the article portion and follow my instructions to process it.}}\\
\makecell[l]{Article:} \\
\makecell[l]{The article begins as follows: \\ \{article\} \\ The article concludes here.} \\

\makecell[l]{Previous summarization:} \\
\makecell[l]{The previous summarization is as follows: \\ \{previous\_sum\} \\ The previous summarization concludes here.} \\

\makecell[l]{Question:} \\
\makecell[l]{\{question\}} \\

\makecell[l]{Instruction:} \\
\makecell[l]{Summarize the partial article to supplement the previous summarization, which can better support the task. } \\
\makecell[l]{If no content needs to be supplemented, please output "null".} \\
\bottomrule
\end{tabular}
\vspace{-10pt}
\end{table}

\section{Limitations and Broad Impact}
\label{sec:limit}
In this paper, we propose LC-Boost, a method dedicated to solving long-context tasks using short contexts. However, there are several limitations we would like to address in the future work:
(1) Although we conduct comprehensive experiments on many tasks and provide theoretical analysis to support our major claim that most long-context tasks are short-context solvable, there may be more complicated scenarios that require understanding the full context in a brute-force setting. LC-Boost might not be able to process such tasks effectively.
(2) As mentioned in Section~\ref{sec:method}, LC-Boost selects actions from a discrete action space. While we argue that the pre-defined action space is versatile enough to handle most scenarios, a more elegant solution would be to predict actions in a continuous space. We conducted preliminary experiments to explore allowing LC-Boost to prompt itself to predict actions without a predefined action space, such as writing prompts or code autonomously. These experiments resulted in highly unstable performance, particularly for models like GPT-3.5, as such requirements are still challenging. We believe that with a much stronger foundation model, LC-Boost could be expected to predict actions in a continuous space.
(3) We choose GPT-3.5 as the foundation model for LC-Boost, instead of open-source LLMs. The reason is that GPT-3.5 is a strong, yet efficient model that can generally understand most instructions. However, we found that most open-source LLMs lack these properties in a zero-shot setting. Fine-tuning these open-source LLMs might be helpful, but constructing such instruction data is infeasible and expensive. 

As discussed in Section~\ref{sec:energy}, LLMs are likely to become a fundamental infrastructure in the near future. At that scale, their energy consumption will pose significant environmental challenges. As shown in Figure~\ref{fig:resource}, LC-Boost avoids processing long contexts directly by decomposing them into shorter contexts. This approach significantly reduces energy consumption as the context length increases, leading to substantial positive environmental impacts. We believe that in the future, more research will focus on green AI initiatives. This paper could serve as an initial spark to inspire further research in this direction, potentially resulting in broader social impact.

\end{document}

%% file: math_command.tex
\usepackage{amsmath,amsfonts,bm}

\def\eqref#1{equation~\ref{#1}}

\def\1{\bm{1}}

\DeclareMathAlphabet{\mathsfit}{\encodingdefault}{\sfdefault}{m}{sl}
\SetMathAlphabet{\mathsfit}{bold}{\encodingdefault}{\sfdefault}{bx}{n}

\def\gA{{\mathcal{A}}}

\def\gX{{\mathcal{X}}}
\def\gY{{\mathcal{Y}}}

\newcommand{\eg}{\textit{e.g.}}

